\pdfoutput=1
\documentclass[11pt]{article}

\usepackage[preprint]{acl}

\usepackage{times}
\usepackage{latexsym}

\usepackage[T1]{fontenc}
\usepackage[utf8]{inputenc}

\usepackage{microtype}
\usepackage{graphicx}
\usepackage{amsmath}
\usepackage{amsthm}
\usepackage{amssymb}
\usepackage{cleveref}
\usepackage{multirow}
\usepackage{booktabs}
\usepackage{pifont}
\newcommand{\cmark}{\ding{51}}%
\newcommand{\xmark}{\ding{55}}%

\usepackage{inconsolata}
\newcommand{\stylesst}{\textsc{SeamlessExpressiveLM}}

\title{\stylesst: Speech Language Model for Expressive Speech-to-Speech Translation with Chain-of-Thought}

\author{Hongyu Gong \qquad Bandhav Veluri \\
  Meta AI \\
  \texttt{\{hygong,bandhav\}@meta.com} \\
}
\begin{document}
\maketitle
% \begin{abstract}
% Expressive speech-to-speech translation (S2ST) preserves semantics and speaker vocal style of source speech, serving a critical role in seamless communication. Recent studies tackle expressive S2ST by leveraging speech language models and various speech representations. Semantic representations mostly encode high-level semantics in speech, and acoustic tokens such as EnCodec capture fine-grained acoustic information. Due to the heterogeneous nature of these representations, existing approaches rely on a cascaded architecture which handles semantic unit translation and acoustic style transfer with different LM components.

% This work proposes \stylesst, a speech language model for expressive S2ST which models both semantic and acoustic tokens in a single model. We adopt chain-to-thought prompting which guides the model to first translate from source to target semantic units and then transfer the acoustic style to target acoustic units. Evaluated on Spanish-to-English and Hungarian-to-English translations, \stylesst~outperforms cascaded LMs in both semantic quality and style transfer, meanwhile achieving better parameter efficiency.
% \end{abstract}

\begin{abstract}
Expressive speech-to-speech translation (S2ST) is a key research topic in seamless communication, which focuses on the preservation of semantics and speaker vocal style in translated speech. Early works synthesized speaker style aligned speech in order to directly learn the mapping from speech to target speech spectrogram.
Without reliance on style aligned data, recent studies leverage the advances of language modeling (LM) and build cascaded LMs on semantic and acoustic tokens. This work proposes \stylesst, a single speech language model for expressive S2ST. We decompose the complex source-to-target speech mapping into intermediate generation steps with \emph{chain-of-thought} prompting. The model is first guided to translate target semantic content and then transfer the speaker style to multi-stream acoustic units. Evaluated on Spanish-to-English and Hungarian-to-English translations, \stylesst~outperforms cascaded LMs in both semantic quality and style transfer, meanwhile achieving better parameter efficiency.
\end{abstract}

\section{Introduction}

% research problem: s2st
Translation is crucial for breaking down language barriers between diverse linguistic backgrounds, and speech-to-speech translation has attracted great interest of research community in fostering cross-lingual spoken communication. Early approaches use encoder-decoder model architecture, and focus on semantic preservation when translating from source to target speech \cite{translatotron,translatotron2,textless-s2st}. The need of seamless communication raises awareness for preserving acoustic information, as speech carries richer information than semantics. For example, the way speech is uttered conveys emotions of the speaker. This motivates the research of S2ST with both semantic and speaker vocal style preserved \cite{seamless}. % Explorations along this line are cascaded S2T

% and preserving speaker vocal style in translation contributes to effective communication across diverse cultural backgrounds.

% existing study: s2st and language model
Language models (LM) with in-context learning capability demonstrate powerful language understanding and generation. The recent progress of discrete speech representations has bridged the gap between continuous speech waveform and discrete text tokens. Semantic tokens such as HuBERT \cite{hubert} and w2v-BERT units \cite{audiopalm} are learned to capture semantic information from speech, and acoustic tokens such as EnCodec \cite{encodec} and SoundStream \cite{soundstream} encode fine-grained acoustic information with multiple codebooks.
The advances in text language model and speech tokens lay the foundation for speech language modeling, which is successfully applied to speech generation tasks \cite{valle}. 
% VALL-E presents controllable text-to-speech synthesis (TTS), transferring the acoustic style of prompt to synthesized speech \cite{valle}. Taking one step further, VALL-E X achieves style preserved S2ST by a cascade of speech-to-text and text-to-speech components \cite{vallex}. 

% challenges
Recent expressive S2ST approaches such as VALL-E X \cite{vallex}, PolyVoice \cite{polyvoice} and AudioPaLM \cite{audiopalm} build a cascade of multiple speech LMs in speech-to-semantic translation and semantic-to-acoustic generation respectively. One LM translates source speech to target semantic tokens such as HuBERT units, and other LMs generates acoustic units conditioned on semantic units and acoustic prompt. We note that multi-LM modeling is faced with common weaknesses of cascaded models such as computational inefficiency and error propagation.

Their design choices of separate LMs for semantic and acoustic generation could be explained by \textbf{the lack of speaker style aligned speech data} for model training. Expressive S2ST model essentially learns the mapping from source to target acoustic units. The end-to-end training would require  speech pairs aligned in both semantics and style \cite{translatotron2}. Despite research efforts in large-scale data mining to create semantically aligned data \cite{speechmatrix}, style aligned speech is in general much less and costly to collect \cite{seamless}. The training of cascaded LMs relaxes the requirements of crosslingual style alignments as the style preservation could be learned from monolingual speech.

\begin{table*}[htbp!]
\caption{A summary of recent S2ST models compared in terms of training data, model architecture and capability.}
\label{tab:model_comparison}
\resizebox{\textwidth}{!}{
\begin{tabular}{ccccccc}
\toprule
 % &  & Enc-Dec S2ST & SeamlessExpressive & Translatatron2 & PolyVoice & \stylesst \\ \midrule\midrule
 &  & \begin{tabular}[c]{@{}c@{}} Enc-Dec S2ST \\ \cite{textless-s2st} \end{tabular} & \begin{tabular}[c]{@{}c@{}} SeamlessExpressive \\ \cite{seamless} \end{tabular} & \begin{tabular}[c]{@{}c@{}} Translatatron2 \\ \cite{translatotron2}\end{tabular} & \begin{tabular}[c]{@{}c@{}} PolyVoice \\ \cite{polyvoice} \end{tabular} & \stylesst \\ \midrule\midrule
\multirow{3}{*}{\begin{tabular}[c]{@{}c@{}}Train\\ data\end{tabular}} & Sem aligned speech & \cmark & \cmark & \cmark & \cmark & \cmark \\
 & Style aligned speech & \xmark & \cmark & \cmark & \xmark & \xmark \\
 & Aligned speech-text & \xmark & \cmark & \cmark & \cmark & \xmark \\ \midrule
\multirow{2}{*}{Model} & Cascaded & \xmark & \cmark & \xmark & \cmark & \xmark \\
 & Components & Encoder-decoder & Encoder-decoder & Encoder-decoder & Three LMs & One LM \\ \midrule
Capability & Style preservation & \xmark & \cmark & \cmark & \cmark & \cmark \\ \bottomrule
\end{tabular}}
\end{table*}

Another challenge resulting in cascaded semantic and acoustic generation is \textbf{the heterogeneity of semantic and acoustic tokens}. The heterogeneity comes from the complimentary information in single-codebook semantic tokens and multi-codebook acoustic tokens. It is challenging to model different types of token generation in a single LM.

% Heterogeneous speech tokens pose challenges to be incorporated into a single model. First of all, semantic and acoustic tokens carry complimentary information. Secondly,  The need of aligned data like in Translatotron2.
% The challenge is that (1) modeling different types of speech tokens in a single model; (2) 

This work proposes \stylesst, a speech LM which models both semantic and acoustic generation for end-to-end expressive S2ST. In this work, we address the challenge of heterogeneous speech tokens by chain-of-thought (CoT) prompting,  and the model adopts multi-step generation from semantic translation towards acoustic translation. As for training data, we only use semantically aligned speech without relying on speaker style aligned data. The trick is to randomly crop the target segment as the acoustic prompt for style preservation.

% our work
% This work proposes \stylesst, a speech language model for direct expressive speech-to-speech translation. We address the heterogeneity of semantic and acoustic tokens the technique of chain-of-thought in model training. 

\Cref{tab:model_comparison} makes a comparison of existing S2ST models and the proposed approach, and our contributions are summarized below:
\begin{itemize}
    \item We propose a novel model to support end-to-end S2ST with speaker style preservation. \stylesst~demonstrates improved translation quality and parameter efficiency compared with the cascaded LMs.
    \item The model is trained with semantically aligned speech, and does not need aligned speech-text data or speaker aligned speech which are required by existing approaches.
    \item We show that semantic and acoustic tokens can be modeled by the same language model despite their heterogeneity.
    \item Our ablation study provides insights into how prompt designs affect speech language model in translation task.
\end{itemize}

\section{Related Work}

\paragraph{S2ST.} The series of Translatotron models have encoder-decoder architecture to translate source speech into target spectrogram, which could be synthesized as waveform with a separately trained vocoder. Translatotron uses a speaker encoder to enable voice conversion in translated speech \cite{translatotron}. Translatotron 2 removes speaker encoder from the model design for the purpose of anti-spoofing, and creates speaker-aligned data with a cross-lingual TTS model for model training \cite{translatotron2}. Therefore the translation model learns to transfer vocal style with a data-driven approach. Translatotron models also leverage textual supervision by using phoneme in the auxiliary task of speech recognition task. 

With semantic units emerging as an efficient semantic representation of speech, unit based translation models are developed. The textless S2ST model learns to map source speech to target units without relying on textual data such as phonemes \cite{textless-s2st}, and the target speech could be synthesized from semantic units with HiFi-GAN vocoder \cite{mlgvocoder}. Despite the good semantic quality, semantic units do not capture speaker vocal style, and thus the unit based S2ST lost the style information in speech translations. To enhance expressivity transfer, it is proposed by \citeauthor{expr-s2st} to integrate speaker and emotion encoder into the translation system. The system consists of speech-to-unit translator and unit-to-speech synthesizer with speaker and emotion embeddings.

%  is a textless model HuBERT units as semantic representations of speech. The textless model does not rely on textual data such as phonemes. Moreover, it simplifies modeling when the target is changed from continuous spectrogram to discrete units. The major limitation is that HuBERT units don’t capture speaker style, and thus the unit based translation lost the style information in translation. 
% \cite{expr-s2st} speech-to-unit and unit-to-speech synthesize with speaker and emotion encoder. 

 % , which is . Data: used phoneme in the auxiliary speech recognition tasks. 
 % Leveraging speaker encoder for voice conversion.  
 % \cite{} . It 

\paragraph{S2ST with language model.} Recent progress in acoustic units such as EnCodec \cite{encodec} and Soundstream \cite{soundstream} capture richer acoustic and style information than semantic speech units by using multiple codebooks to encode residual information in each codebook. This also unlocked the speech language modeling in vocal style transfer \cite{valle}. 

With acoustic units as translation target, the translation model is trained to preserve the style of source speaker. 
Existing works are built upon a cascaded architecture \cite{polyvoice,vallex} consisting of speech-to-unit translation, primary acoustic unit and residual acoustic unit generation. The first component translates target semantic units from source speech. The other two components take care of speaker style transfer, and generate the first stream of acoustic units and remaining streams sequentially. This type of model has general limitations of cascaded architecture, which are inefficiency and error propagation. A multi-task learning framework is recently designed \cite{peng2024mslm}, which improves model efficiency by sharing parameters between the first and second components. However, it still goes with cascaded training and inference of the three components.

We come up with \stylesst~which is trained in an end-to-end manner with chain-of-thought. It shows better parameter efficiency and improved performance compared with the cascaded model.
 
\begin{figure*}[htbp!]
    \centering
    \includegraphics[width=1.0\linewidth]{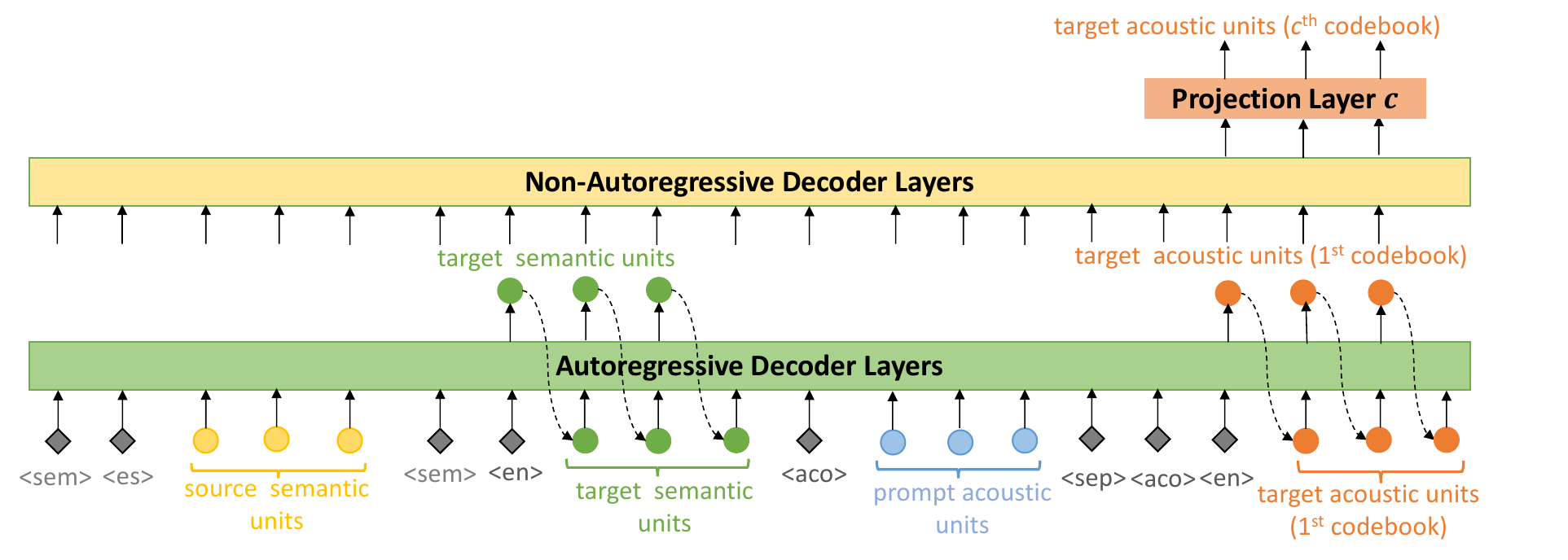}
    \caption{Model architecture of \stylesst.}
    \label{fig:model}
\end{figure*}

\section{Model}

This section introduces \stylesst, a decoder-only language model for style transferred speech-to-speech translation.

\subsection{Speech Tokenizers}

Speech tokenizers convert continuous speech waveform into a sequence of discrete units. HuBERT is used to derive semantic units of speech which mainly keep the semantic information \cite{hubert}. EnCodec extracts multi-codebook units to carry more fine-grained acoustic information in speech such as speaker vocal style and intonation\cite{encodec}. Suppose that EnCodec has $C$ codebooks, and $C=8$ in our experiments. Both HuBERT and EnCodec units are leveraged in our modeling as described below.

\subsection{Architecture}

We define notations here to simplify the discussion in the rest of this paper. Source semantic units are denoted as $\mathbf{S}_{\text{src}}$, target semantic units are $\mathbf{S}_{\text{tgt}}$, prompt acoustic units are $\mathbf{A}_{\text{prompt}}^{\{c: 1\leq c\leq C\}}$ and target acoustic units are $\mathbf{A}_{\text{tgt}}$. Without speaker style aligned data, expressive S2ST is formulated to learn the mapping: $\mathbf{S}_{\text{src}}$ + $\mathbf{A}_{\text{prompt}}$ $\rightarrow$ $\mathbf{A}_{\text{tgt}}$, where $\mathbf{S}_{\text{src}}$ controls the target semantics and $\mathbf{A}_{\text{prompt}}$ influences the target style.

\stylesst~adopts chain-of-thought prompting, learning the complex mapping through multi-step reasoning: 
\begin{align}
\label{eq:chain}
% ($\mathbf{S}_{\text{src}}$) $\rightarrow$ $\mathbf{S}_{\text{tgt}}$ (+ $\mathbf{A}_{\text{prompt}}$) $\rightarrow$ $\mathbf{A}_{\text{tgt},1}$ $\rightarrow$ $\mathbf{A}_{\text{tgt},c}$ ($c>1$). 
(\mathbf{S}_{\text{src}}) \rightarrow \mathbf{S}_{\text{tgt}} + ( \mathbf{A}_{\text{prompt}}) \rightarrow \mathbf{A}_{\text{tgt},1} \rightarrow \mathbf{A}_{\text{tgt},c>1},
\end{align}
where each $\rightarrow$ is one reasoning step, and variables in $(\cdot)$ are given while other variables $\mathbf{S}_{\text{tgt}}$, $\mathbf{A}_{\text{tgt},1}$ and $\mathbf{A}_{\text{tgt},c}$ are predicted by the model.

As illustrated in \Cref{fig:model}, the input sequence reflects the chain of thoughts in translation process. The model starts from semantic translation, proceeds to transfer the style from prompt acoustic units to the first acoustic stream and lastly learns the mapping to residual acoustic streams.

% As shown in \Cref{fig:model}, the model takes semantic and multi-stream acoustic tokens of source speech, which are denoted as $\mathbf{S}_{\text{src}}$ and $\mathbf{A}_{\text{src}}$ respectively. The target output is acoustic tokens $\mathbf{A}_{\text{tgt}}$ in target language which carries the same semantic meaning and vocal style of source speech.

\paragraph{Embedding layer.} We construct embedding tables for semantic and acoustic units to vectorize the speech tokens before passing them to the speech language model. 
For semantic units and first-stream acoustic units, their embeddings can be retrieved directly from the embedding table $\text{Emb}(\cdot)$. As for prompt acoustic tokens from multiple streams, we sum up embeddings from all codebooks as the acoustic embedding in each position.  
\begin{align}
    \text{Emb}(\mathbf{A}_{\text{prompt}}^{i}) = \sum\limits_{c=1}^{C}\text{Emb}(\mathbf{A}_{\text{prompt},c}^{i})
\end{align}

\paragraph{Autoregressive (AR) Layers.}
% Our proposed model is built upon an \emph{autoregressive} (AR) Transformer decoder and a \emph{non-autoregressive} (NAR) decoder. 
\Cref{fig:model} demonstrates the overall model architecture. The bottom layers of \stylesst~are AR Transformer decoder layers, which learn the sequence modeling of semantic units and primary acoustic units. In AR layer, each token only attends to its previous tokens and the sequential order of tokens is preserved. Suppose that AR decoder outputs are $\mathbf{H}_{i}$ for $i$-th token in the sequence, which is projected by matrix $\mathbf{W}^{\text{ar}}$ to a probability distribution over the unit vocabulary.

The model is prompted to generate semantic and acoustic units progressively. The semantic units of source speech $\mathbf{S}_{\text{src}}$ are first provided as the semantic prompt, and the model estimates the probability of semantic units of target speech. 
\begin{align}
    \tilde{\mathbf{P}}(\mathbf{S}_{\text{tgt}}^{i}|\mathbf{S}_{\text{src}}, \mathbf{S}_{\text{tgt}}^{<i}) = \text{softmax}\left(\mathbf{W}^{\text{ar}}\mathbf{H}^{\text{ar}}_{i}\right)
\end{align}

Moving forward from semantic translation, the model is further provided with the acoustic units of source speech $\mathbf{A}_{\text{src}}$ as acoustic prompt, and instructed to predict the target acoustic units in the first codebook, $\mathbf{A}_{\text{tgt},1}$.

 % of the first acoustic codebook. AR decoder is trained to predict the first stream of target acoustic units .

\begin{align}
    \nonumber
    &\tilde{\mathbf{P}}(\mathbf{A}_{\text{tgt}, 1}^{j}|\mathbf{S}_\text{src}, \mathbf{S}_\text{tgt}, \mathbf{A}_\text{prompt}, \mathbf{A}_{\text{tgt}, 1}^{<j}) \\
    &= \text{softmax}\left(\mathbf{W}^{\text{ar}}\mathbf{H}_{j}\right),
\end{align}
where the acoustic unit is the $j$-th token in the sequence.

\paragraph{Non-Autoregressive (NAR) Layer.} On top of AR layers which model semantic and first-stream acoustic unit sequence, we have NAR layers to model acoustic unit sequence of remaining codebooks $\mathbf{A}_{\text{tgt},c}~(c>1)$. The design of NAR layers is inspired by VALL-E \cite{valle}, which enable efficient decoding of multi-stream acoustic units. We note that there are \textbf{key differences} in the design. VALL-E \emph{sequentially} decodes each stream by conditioning the current prediction on previous streams of units. In our model, we support \emph{parallel decoding} of streams in NAR component.

The decoder outputs $\mathbf{H}^{\text{ar}}$ of AR layers is routed to the NAR layer. In NAR layers, each token could attend to the whole sequence to access broader context. To distinguish multiple acoustic streams, we use different projection matrices $\{\mathbf{W}_{c}\}$ for each codebook $c$ $(c>1)$. The output of NAR layer, $\mathbf{H}^{\text{nar}}_{i}$ at position $i$, is mapped to the vocabulary of the $c$-th acoustic codebook:

\begin{align}
    \nonumber
   &\tilde{\mathbf{P}}(\mathbf{A}_{\text{tgt},c}^{j} | \mathbf{S}_{\text{src}}, \mathbf{S}_{\text{tgt}}, \mathbf{A}_{\text{prompt}}, \mathbf{A}_{\text{tgt},1}) \\
   &=\text{softmax}\left(\mathbf{W}^{\text{nar}}_{c}\mathbf{H}^{\text{nar}}_{j}\right)
\end{align}

\subsection{Training}
\label{subsec:model:training}

\textbf{Acoustic prompt.} Prompt acoustic units $\mathbf{A}_{\text{prompt}}$ provide all acoustic information to be preserved in the target speech. Ideally given source and target speech which both semantically and acoustically aligned, the model could simply use source units as the acoustic prompt. Now we want to relax the data requirements and use only semantically aligned speech. We carve a portion of target speech as the acoustic prompt. To prevent the model from naively copy-paste the acoustic prompt in target acoustic generation, the acoustic prompt is randomly selected at each train step and the prompt length has non-trivial effect on mdel performance. The designs of acoustic prompt is discussed in \Cref{subsec:ablation:acoustic}.

\paragraph{AR loss.} The autoregressive layers takes care of target semantic units and first-stream acoustic units. We use chain-of-thought to train the model to first translate speech semantically and then transfer speaker vocal style to the target speech. AR layers of \stylesst~are trained with next token prediction similar to existing language models. It is noted that source semantic units and source acoustic units are given as the prompt, and these tokens do not count towards the training loss. Assume that $\mathbf{P}(\cdot)$ is the ground truth probability (either 0 or 1) of target units.

\begin{align}
    \nonumber
    &\mathbb{L}_{\text{ar}} = - \sum\limits_{i}\mathbf{P}(\mathbf{S}_{\text{tgt}})\log\tilde{\mathbf{P}}(\mathbf{S}_\text{{tgt}}|\mathbf{S}_{\text{src}}) - \\
    &\sum\limits_{j}\mathbf{P}(\mathbf{A}_{\text{tgt}, 1}^{j})\log \tilde{\mathbf{P}}(\mathbf{A}_{\text{tgt}, 1}^{j}|\mathbf{S}_\text{src}, \mathbf{S}_\text{tgt}, \mathbf{A}_\text{prompt}, \mathbf{A}_{\text{tgt}, 1}^{<j})
\end{align}

\paragraph{NAR loss.} In each train step, we randomly select one stream $c$ of acoustic units for NAR layers to predict.
% Remaining streams of acoustic units are predicted from NAR layers, and cross-entropy loss is applied to the unit prediction of each stream. 
\begin{align}
    \nonumber
    &\mathbb{L}_{\text{nar}} = 
    -\sum\limits_{j}\mathbf{P}(\mathbf{A}_{\text{tgt},c}^{j})\cdot \\ 
    &\log\tilde{\mathbf{P}}(\mathbf{A}_{\text{tgt},c}^{j} | \mathbf{S}_{\text{src}}, \mathbf{S}_{\text{tgt}}, \mathbf{A}_{\text{prompt}}, \mathbf{A}_{\text{tgt},1})
\end{align}

The total loss is $\mathbb{L}=\mathbb{L}_{\text{ar}} + \mathbb{L}_{\text{nar}}$, and \stylesst~is trained in an end-to-end manner.

% \textbf{Finetuning.} Further finetuning to predict target acoustic units directly from source?

\subsection{Inference}
\label{subsec:model:inference}

The model decodes target semantic units with beam search, and then generates acoustic units with temperature sampling. Target speech is synthesized by EnCodec decoder from predicted acoustic units.

\section{Experiments}

\begin{table}[htbp!]
\begin{tabular}{cccc}
\toprule
 & Samples (k) \# & Avg (s) & Total (h) \\ \midrule
Es-En & 250 / 1.9 & 7.2 / 10.0 & 500 / 5.4 \\
%  & Eval & 1,947 & 10.0 & 5.4 \\
Hu-En & 300 / 1.0 & 9.2 / 13.8 & 763 / 3.9 \\
% Eval2 & 1,816 & 10.1 & 5.1 \\ 
\bottomrule
\end{tabular}
\caption{Statistics of training and evaluation data including sample size, average duration in seconds and total duration in hours. The first value is training statistics, and the second value is test statistics.}
\label{tab:data}
\end{table}

This section covers empirical details of model training and evaluation on the task of speech-to-speech translation with speaker style transfer. We consider \emph{Spanish-to-English} (Es-En) and \emph{Hungarian-to-English} (Hu-En) translation as representative translations between similar and distant languages.

\begin{table*}[htbp!]
\centering
\resizebox{\textwidth}{!}{\begin{tabular}{ccccccc}
\toprule
\multirow{2}{*}{} & \multirow{2}{*}{\begin{tabular}[c]{@{}c@{}}Semantic\\ Params\end{tabular}} & \multirow{2}{*}{\begin{tabular}[c]{@{}c@{}}Acoustic\\ Params\end{tabular}} & \multicolumn{2}{c}{Es-En} & \multicolumn{2}{c}{Hu-En} \\
 &  &  & ASR-BLEU ($\uparrow$) & VSim ($\uparrow$) & ASR-BLEU ($\uparrow$) & VSim ($\uparrow$) \\ \midrule
\multicolumn{1}{l}{Enc-Dec} & 187M & - & 20.05 & 0.024 & 10.43 & 0.070 \\
\multicolumn{1}{l}{Cascaded LMs} & 155M & 314M & 16.86 & 0.292 & 9.16 & 0.402 \\ \midrule\midrule
\multicolumn{1}{l}{\stylesst} & \multicolumn{2}{c}{312M} & 17.02 & 0.327 & 9.73 & 0.431 \\
\multicolumn{1}{l}{- no chain-of-thought} & \multicolumn{2}{c}{312M} & 7.20 & 0.392 & 5.35 & 0.486 \\
\multicolumn{1}{l}{- no semantic prompt} & \multicolumn{2}{c}{312M} & 6.41 & 0.548 & 4.41 & 0.543 \\
\multicolumn{1}{l}{- no acoustic prompt} & \multicolumn{2}{c}{312M} & 20.16 & 0.024 & 10.55 & 0.067 \\ 
\bottomrule
\end{tabular}}
\caption{Spanish-English and Hungarian-English translation results including model parameter size, ASR-BLEU and vocal style similarity.}
\label{tab:results}
\end{table*}

\paragraph{Data.} We used in-house semantically aligned data consisting of $250$k Spanish-English and $300$k Hungarian-English speech pairs. A set of $1$k samples were randomly taken from training data as the validation set.
% There are $250$k speech pairs with a total duration of $500$ hours. A small set of $5$k samples was taken as validation set, and others serve as training data.
% To evaluate Spanish-to-English translation, we prepare two evaluation sets with $1,947$ ad $1,816$ samples respectively. 
\Cref{tab:data} summarizes all data statistics.

\paragraph{Automatic evaluation metrics.} We reuse \emph{ASR-BLEU} metric to measure the semantic translation quality as existing studies of speech translation. Generated audios are first transcribed with ASR tools into texts, and we use the medium-sized Whisper English ASR model\footnote{https://huggingface.co/openai/whisper-medium.en} in this work. BLEU evaluates the ngram overlap between the transcripts and ground truth translations.

Another important metric is \emph{vocal style similarity} (VSim), quantifying how similar the generated audios sound in comparison with the source speech in term of vocal style. Following \cite{seamless}, we extract vocal style embeddings of generated and source speech with a pretrained WavLM encoder \cite{wavlm} and compute the cosine similarity of their embeddings. Higher embedding similarity indicates better vocal style transfer. Furthermore we report \emph{model size}, the number of parameters, to reflect the model efficiency. 
% As for inference speed, we measure the \emph{decoding speed}, i.e., the number of decoded tokens per second on a single GPU.

\begin{table}[htbp!]
\centering
\begin{tabular}{ccc}
\toprule
 & Es-En & Hu-En \\ \midrule
Cascaded LMs & 3.34 & 3.56 \\
\stylesst & 3.28 & 3.82 \\
\bottomrule
\end{tabular}
\caption{MOS of model translations.}
\label{tab:mos}
\end{table}

\paragraph{Subjective metrics.} Besides automatic metrics, we include Mean Opinion Score (MOS) as the subjective metric to measure the speech quality with scores ranging from $1$ to $5$ (higher score indicates better quality). We get $50$ samples from the model outputs in each test set, and two annotators independently evaluate the quality, and we report the score averaged over annotators and test samples.

% \begin{table}[htbp!]
% \centering
% \caption{The effect of prompt ratio in \stylesst training.}
% \label{tab:prompt_ratio}
% \begin{tabular}{ccc}
% \toprule
% \multirow{2}{*}{\begin{tabular}[c]{@{}c@{}}Acoustic\\ prompt ratio\end{tabular}} & \multicolumn{2}{c}{Eval1} \\
%  & ASR-BLEU ($\uparrow$) & VSim ($\uparrow$) \\ \midrule\midrule
% {[}0.15, 0.20) & x & x \\
% {[}0.20, 0.25) & x & x \\
% {[}0.25, 0.30) & x & x \\
% {[}0.45, 0.55) & x & x \\ \bottomrule
% \end{tabular}
% \end{table}

\paragraph{Models.} We included recent S2ST models which are only trained on semantically aligned speech data for a fair comparison with \stylesst.
% Recent studies leveraging speech language model for S2ST are cascaded models, and we include them as baselines below.
\begin{itemize}
    \item Enc-Dec S2ST. We included a textless translation model with encoder-decoder architecture as a strong baseline for semantic translation \cite{speechmatrix}. The source speech is encoded by  convolutional and Transformer encoder layers, and its decoder predicts the target semantic units. A separately trained vocoder synthesizes target speech from semantic units.
    \item Cascaded LMs\footnote{We use the architecture of PolyVoice, and replace SoundStream with Encodec as acousitc units}. PolyVoice \cite{polyvoice} tackles S2ST with a cascade of three LMs for (1) semantic unit translation, (2) the first acoustic stream generation, (3) remaining acoustic streams generation. LM-1 is a decoder-only model translating source to target semantic units, LM-2 and LM-3 follow VALL-E for acoustic generation which is trained on target speech i.e., English audios.
    % \item Two-stage cascaded S2ST. An optimization of three-stage approach is to merge the first two stages, and a single model is trained with multi-tasks covering both semantic unit and first-stream acoustic unit generation. We note that this efficiently reduces the model size, but its inference time stays the same as three-stage approach the three tasks are executed sequentially.
    % \item Semantic-only LM.
\end{itemize}

\stylesst~consists of $12$ AR decoder layers and $12$ NAR layers. The token embedding dimension is $512$. Each layer is Transformer decoder layer with $16$ attention heads, the layer dimension of $1024$ and the feedforward dimension of $4096$. The model is trained with $0.1$ dropout at the learning rate of $0.0002$. The cascaded model consists of three LMs each of which has $12$ Transformer decoder layers and other hyperparameters same as \stylesst. As for Enc-Dec S2ST, we set its convolution layers as $2$, encoder layers of $6$ and decoder layers of $6$ with layer and feedforward dimensions of $1024$ and $4096$ so that its parameter size match that of semantic LM in the cascaded model.

% train and inference
\paragraph{Training.} For \stylesst, the training acoustic prompt ratio is uniformly sampled between $0.25$ and $0.3$, and the inference prompt ratio is $0.3$ in Spanish-to-English translation. The training ratio is $0.20$-$0.25$, and the inference ratio is $0.2$ for Hungarian-to-English translation. \Cref{subsec:ablation:acoustic} discusses how the training and inference ratios are selected for Spanish-to-English. For cascaded LMs, the acoustic prompt is fixed as $3$ seconds as used in \cite{valle}. 

% \paragraph{Inference.} We set beam search size of $10$ during inference. The model decodes target semantic unit and then sample acoustic units with the temperature of $0.9$. 
\paragraph{Inference.} We set beam search size of $10$ to all models during inference. The \stylesst samples acoustic units with the temperature of $0.9$. For cascaded LMs, its semantic-to-acoustic generation uses a temperature of $0.8$ following \cite{valle}. For LM-based S2ST models, target speech is synthesized by EnCodec decoder from acoustic units. For Enc-Dec, since it predicts semantic units, a pretrained unit based HiFi-GAN vocoder is applied to synthesize speech following \cite{textless-s2st}.

% \paragraph{Results.} \Cref{tab:results} reports model size, ASR-BLEU and VSim of different S2ST models for Es-En and Hu-En. Enc-Dec model serves as the upper bound of semantic translation  quality, while it does not preserve speaker vocal style. \stylesst~outperforms cascaded LMs with $+12\%$ similarity in Es-En test set. It performs on par with Cascaded LMs in terms of semantic translation as reflected by comparable ASR-BLEU scores.  In \Cref{tab:mos}, \stylesst~shows good MOS in comparison with cascaded LMs. % demonstrating good audio quality.}
%As shown in \Cref{tab:mos}, \textcolor{red}{MOS of \stylesst~is comparable to cascaded LMs, demonstrating good audio quality.} 
% We note that \stylesst~has better efficiency with $312$M parameters in comparison with Cascaded LMs with $469$M parameters. 
% It demonstrates that end-to-end training of \stylesst improves the translation performance .

\begin{figure*}
    \centering
    \includegraphics[width=0.55\linewidth]{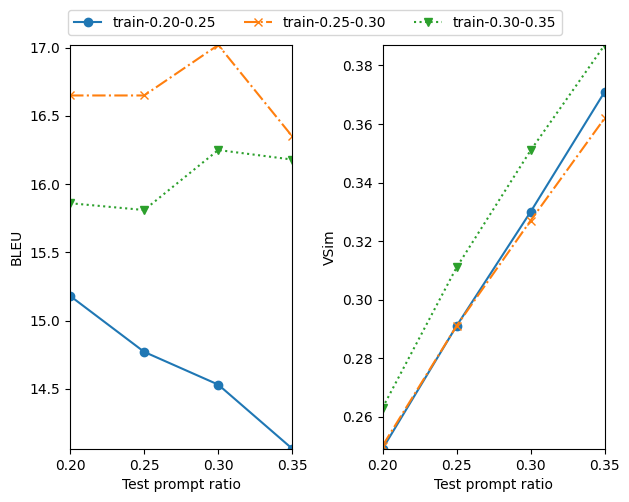}
    \caption{An ablation study of acoustic prompt ratios in Es-En training and inference.}
    \label{fig:prompt_ratio}
\end{figure*}

\subsection{Semantic Prompt}
\label{subsec:ablation:semantic}

\paragraph{Results.} \Cref{tab:results} reports model size, ASR-BLEU and VSim of different S2ST models for Es-En and Hu-En. Enc-Dec model serves as the upper bound of semantic translation quality, while it does not preserve speaker vocal style. \stylesst~outperforms cascaded LMs with $+10.7\%$ and $+7.2\%$ similarity in two directions. It performs on par with Cascaded LMs in terms of semantic translation as reflected by comparable ASR-BLEU scores. In \Cref{tab:mos}, \stylesst~shows good MOS in comparison with cascaded LMs. 
We note that \stylesst~has better efficiency with $312$M parameters in comparison with Cascaded LMs with $469$M parameters.

% Enc-Dec S2ST and the cascaded LMs have achieve comparable ASR-BLEU, as they have similar parameter size for semantic translation. Comparing \stylesst and cascaded LMs, \stylesst with $312$M parameters achieves +$??$ ASR-BLEU and comparable VSim than cascaded LMs with $469$M parameters in Eval1, and ASR-BLEU and VSim in Eval2. 

\section{Ablation Study}
% We now perform ablation study to provide more insights into the model. 
Semantic units and prompt acoustic units make a chain of S2ST prompts for \stylesst. A natural question to ask is how effective such prompt design is for speech LM training. Therefore we tried other prompting strategies as the ablation study. 

\subsection{Chain-of-Thought Prompting}
\label{subsec:ablation:cot}

As shown in \Cref{eq:chain}, CoT constructs multi-step reasoning towards the target $\mathbf{A}_{\text{tgt}}$ given the training samples $\{(\mathbf{S}_{\text{src}}, \mathbf{S}_{\text{tgt}}, \mathbf{A}_{\text{prompt}})\}$. In an ablation study of the effectiveness of chain-of-thought, we design a variant strategy without CoT by breaking chain-of-thought into two prompting tasks: \\
\noindent(1) $\mathbf{S}_\text{src}+\mathbf{A}_\text{prompt} \rightarrow \mathbf{A}_\text{tgt}$: this is basically the expressive translation task predicting target acoustic units conditioned on source semantic units and prompt acoustic units; \\
\noindent(2) $\mathbf{S}_\text{tgt}+\mathbf{A}_\text{prompt} \rightarrow \mathbf{A}_\text{tgt}$: this is a monolingual style transfer task predicting target acoustic units conditioned on target semantic units and prompt acoustics. 

This strategy provides essentially the same information as CoT prompting does for model training. It uses multi-task learning in place of multi-step reasoning. In \Cref{tab:results}, the row ``no chain-of-thought'' shows a drop of $9.82$ and $4.38$ in ASR-BLEU for Es-En and Hu-En respectively. It suggests that CoT helps the model with better semantic preservation in translation process.

% The first strategy removes CoT by breaking chain-of-thought into two prompting tasks: (1) $\mathbf{S}_\text{src}+\mathbf{A}_\text{prompt} \rightarrow \mathbf{A}_\text{tgt}$: predicting target acoustic units conditioned on source semantic units and prompt acoustic units; (2) $\mathbf{S}_\text{tgt}+\mathbf{A}_\text{prompt} \rightarrow \mathbf{A}_\text{tgt}$: predicting target acoustic units conditioned on target semantic units and prompt acoustics. The two prompts contain essentially the same information as the CoT prompt. In \Cref{tab:results}, the row ``no chain-of-thought'' shows a drop of $9.82$ and $4.38$ in ASR-BLEU for Es-En and Hu-En respectively. It suggests that CoT helps the model with better semantic preservation in translation process.

To quantify the importance of semantic prompt in modeling, we experiment with another strategy by removing target semantic units from CoT prompting. Specifically, the model is trained to directly predict target acoustic units conditioned on source semantic units and prompt acoustic units. The row of ``no semantic prompt'' in \Cref{tab:results} shows semantic degradation with a drop of $10.61$ ASR-BLEU in Es-En and $5.32$ in Hu-En, suggesting that semantic prompt plays a critical role in providing semantic cues for S2ST modeling.

\subsubsection{Semantic Prompt}
\label{subsec:ablation:semantic}

To quantify the importance of semantic prompt in modeling, we experiment with another strategy by removing target semantic units from CoT prompting. The model is trained to directly predict target acoustic units conditioned on source semantic units and prompt acoustic units, i.e., $\mathbf{S}_\text{src}+\mathbf{A}_\text{prompt} \rightarrow \mathbf{A}_\text{tgt}$.

The row of ``no semantic prompt'' in \Cref{tab:results} shows semantic degradation with a drop of $10.61$ ASR-BLEU in Es-En and $5.32$ in Hu-En, suggesting that semantic prompt plays a critical role in providing semantic cues for S2ST modeling.

\subsection{Acoustic Prompt}
\label{subsec:ablation:acoustic}

Similar to the ablation study of semantic prompt, we remove acoustic prompt to evaluate its effect on the translation model. We use the trained \stylesst~to generate target semantic units conditioned on source semantics (i.e., $\mathbf{S}_\text{src} \rightarrow \mathbf{S}_\text{tgt}$), and synthesize speech from target units with the vocoder that is used by Enc-Dec model. The speech generated without acoustic prompt does not carry source vocal style, as indicated by the low VSim score from the row ``no acousti prompt'' in \Cref{tab:results}. On the other hand, it achieves the best ASR-BLEU among all models, showing that decoder-only speech LM is as strong as encoder-decoder architecture in semantic translation.

Moreover as a portion of target speech is taken as the acoustic prompt in \stylesst~training, one important hyperparamter in this work is the prompt ratio. Taking Spanish-to-English as an example, we train multiple models with three sets of prompt ratio ranges: ($0.20$, $0.25$), ($0.25$, $0.3$) and (0.30, 0.35). For each train sample, a prompt ratio is uniformly selected from the given range. As for inference, we apply different prompt ratios to test samples, and measure how ASR-BLEU and VSim change with it.

As shown in \Cref{fig:prompt_ratio}, the training prompt range ($0.25$, $0.30$) achieves the highest ASR-BLEU with test prompt ratio of $0.3$. Short acoustic prompt cannot provide sufficient acoustic information for the translation, while long acoustic prompt might encourage the model to copy and paste the prompt as it is taken from the target speech. 
Models trained with ($0.25$, $0.30$) and ($0.30$, $0.35$) achieve the best ASR-BLEU when the test prompt ratio is set as $0.30$. For model trained with ($0.20$, $0.25$), its ASR-BLEU drops when test prompt ratio increases. We can see consistent improvement of VSim with increased test prompt in all three models.

\section{Conclusion}

We propose \stylesst~to achieve expressive S2ST with chain-of-thought, which unifies semantic and acoustic language modeling, and improves efficiency over cascaded approach.

\section{Limitations and Risks}

\paragraph{Limitations.} One limitation of this work is that it focuses on speech-only modeling. Existing works have shown improved speech translation quality with different data types such as aligned speech-text data. The experiments are conducted in a limited setting where the model and training data sizes are not large. Scaling up model and data would help to enhance the translation performance. 

\paragraph{Ethical considerations.} The proposed model is intended to use for expressive speech-to-speech translation with both semantics and vocal style preserved. In case of failure, the system will generate inaccurate semantic meaning or vocal styles, thus negatively impacting the translation quality. One example of unintended use is that bad actors misappropriate the model for nefarious activities such as online scams.

\bibliography{custom}

\end{document}